\pdfoutput=1

\documentclass[11pt]{article}

\usepackage[]{acl}

\usepackage{times}
\usepackage{latexsym}
\usepackage{booktabs}
\usepackage{graphicx}
\usepackage{subcaption}
\usepackage{amsmath, amssymb, amsthm}
\usepackage{float}

\usepackage[T1]{fontenc}

\usepackage[utf8]{inputenc}

\usepackage{microtype}

\title{Understanding Game-Playing Agents with Natural Language Annotations}

\author{Nicholas Tomlin \qquad Andre He \qquad Dan Klein \\
  Computer Science Division, University of California, Berkeley \\
  \texttt{\{nicholas\_tomlin, andre.he, klein\}@berkeley.edu}}

\date{}

\begin{document}
\maketitle
\begin{abstract}
We present a new dataset containing 10K human-annotated games of Go and show how these natural language annotations can be used as a tool for model interpretability.
Given a board state and its associated comment, our approach uses linear probing to predict mentions of domain-specific terms (e.g., \textit{ko}, \textit{atari}) from the intermediate state representations of game-playing agents like AlphaGo Zero. We find these game concepts are nontrivially encoded in two distinct policy networks, one trained via imitation learning and another trained via reinforcement learning. Furthermore, mentions of domain-specific terms are most easily predicted from the later layers of both models, suggesting that these policy networks encode high-level abstractions similar to those used in the natural language annotations.
\end{abstract}

\section{Introduction}

\begin{figure}[h]
\centering
\includegraphics[width=\linewidth]{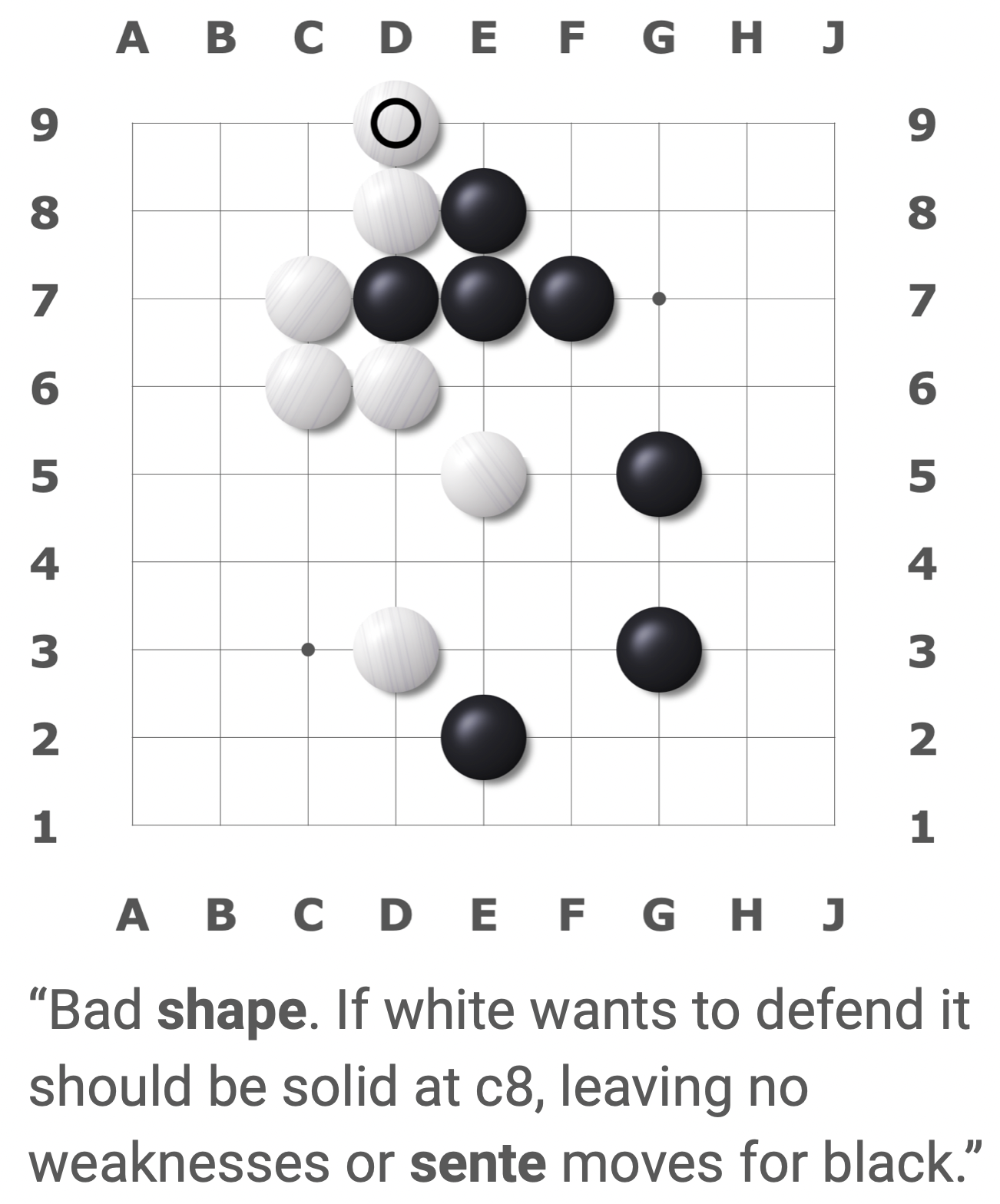}
\caption{Example comment from our dataset, with domain-specific keywords (\textit{shape}, \textit{sente}) highlighted. Although this comment is from a 9 $\times$ 9 game for illustrative purposes, our dataset primarily focuses on annotations from 19 $\times$ 19 games.}
\label{fig:comment}
\end{figure}

\begin{figure*}[h]
\centering
\begin{subfigure}{.2\textwidth}
  \centering
  \includegraphics[width=\linewidth]{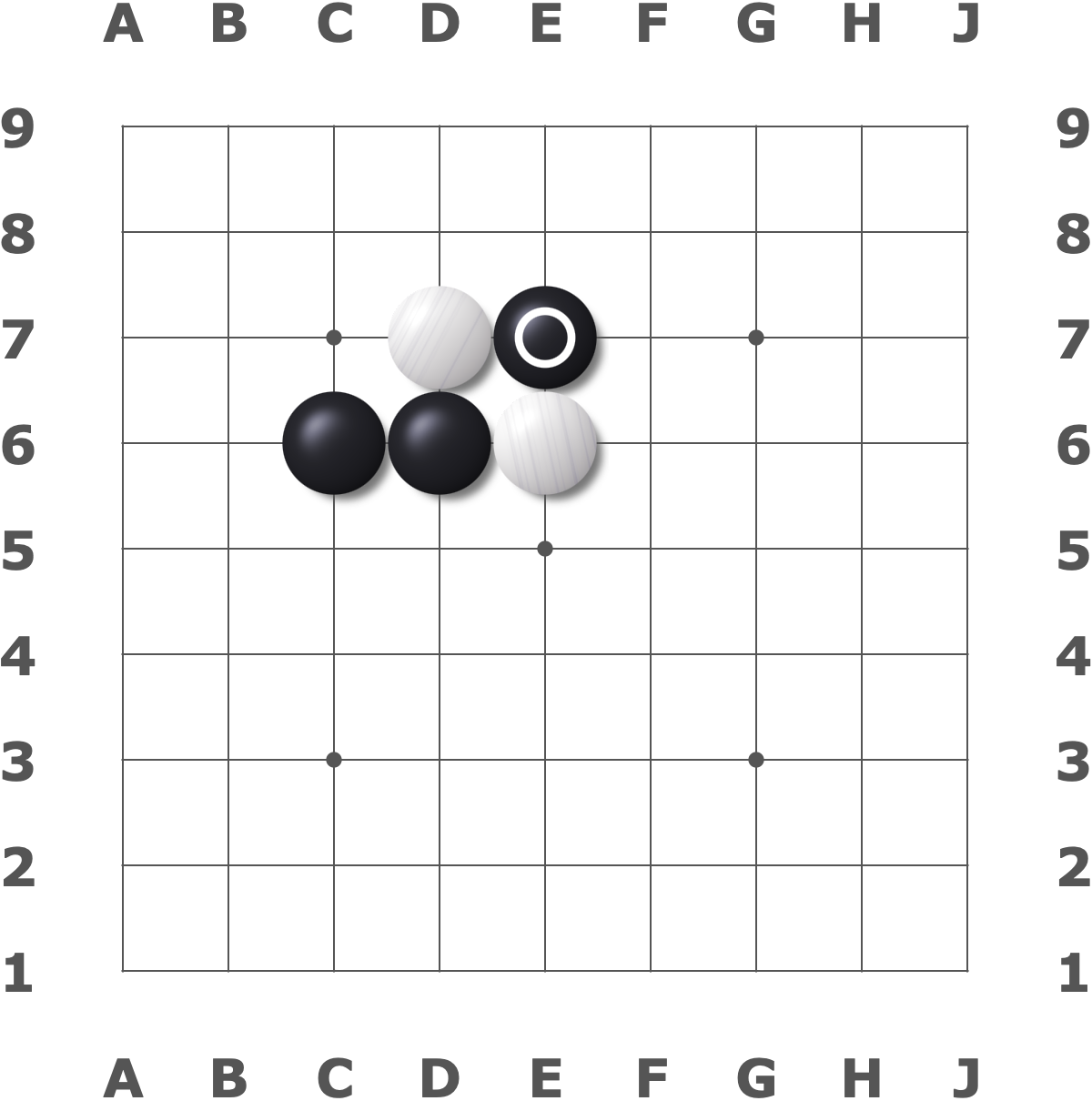}
  \caption{Cut}
  \label{fig:cut}
\end{subfigure}\hfill
\begin{subfigure}{.2\textwidth}
  \centering
  \includegraphics[width=\linewidth]{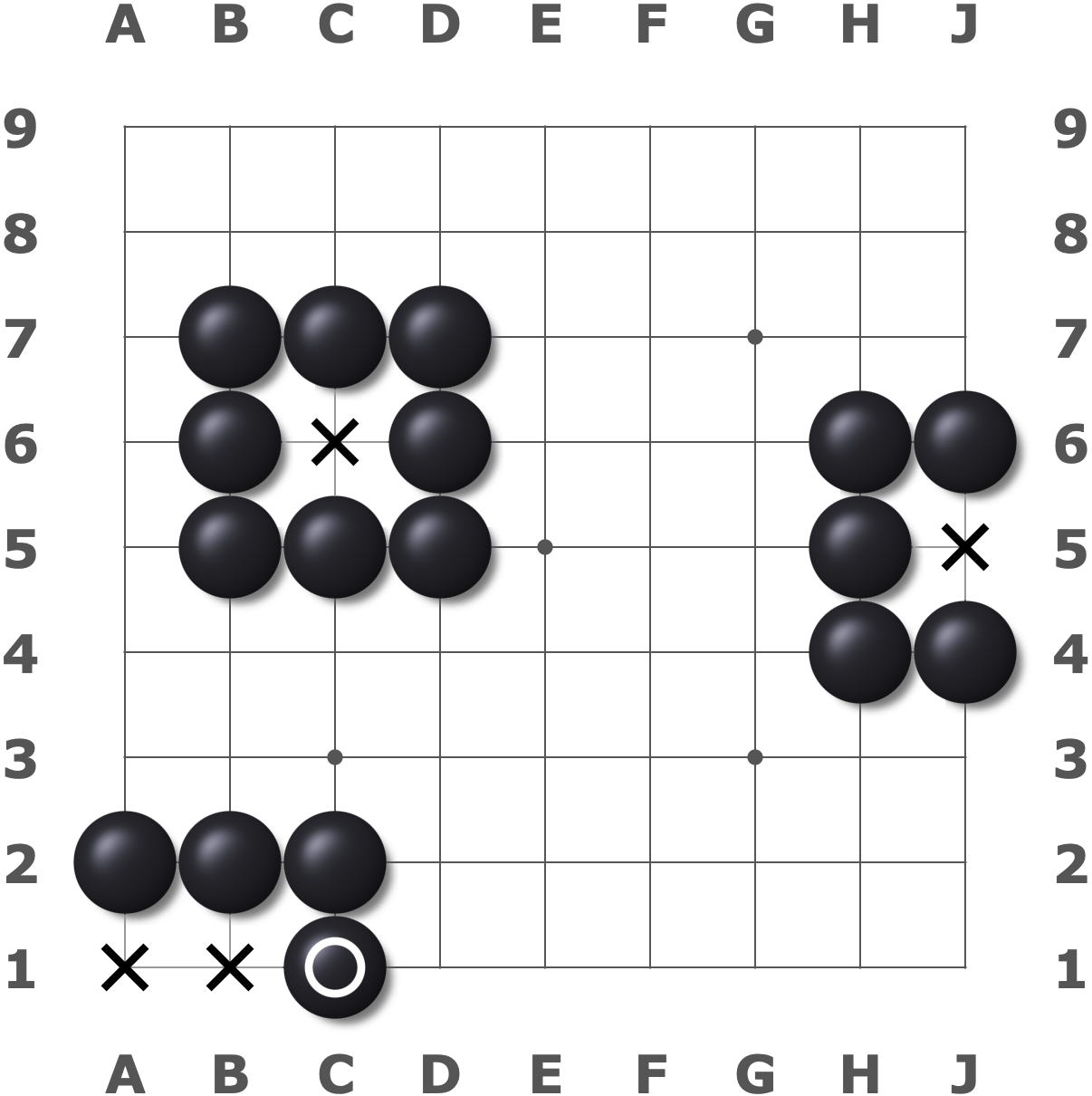}
  \caption{Eyes}
  \label{fig:eye}
\end{subfigure}\hfill
\begin{subfigure}{.2\textwidth}
  \centering
  \includegraphics[width=\linewidth]{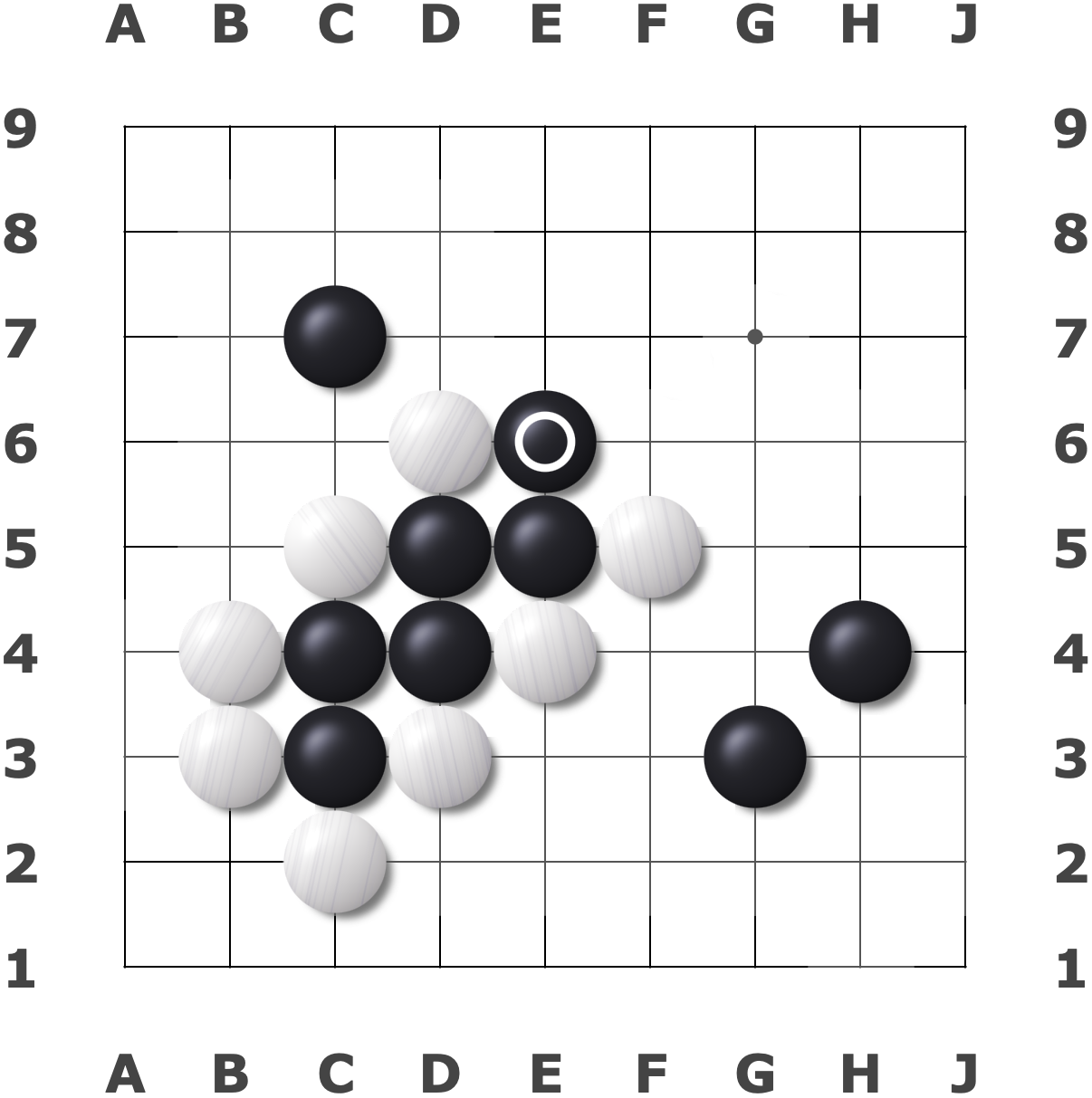}
  \caption{Ladder}
  \label{fig:ladder}
\end{subfigure}\hfill
\begin{subfigure}{.2\textwidth}
  \centering
  \includegraphics[width=\linewidth]{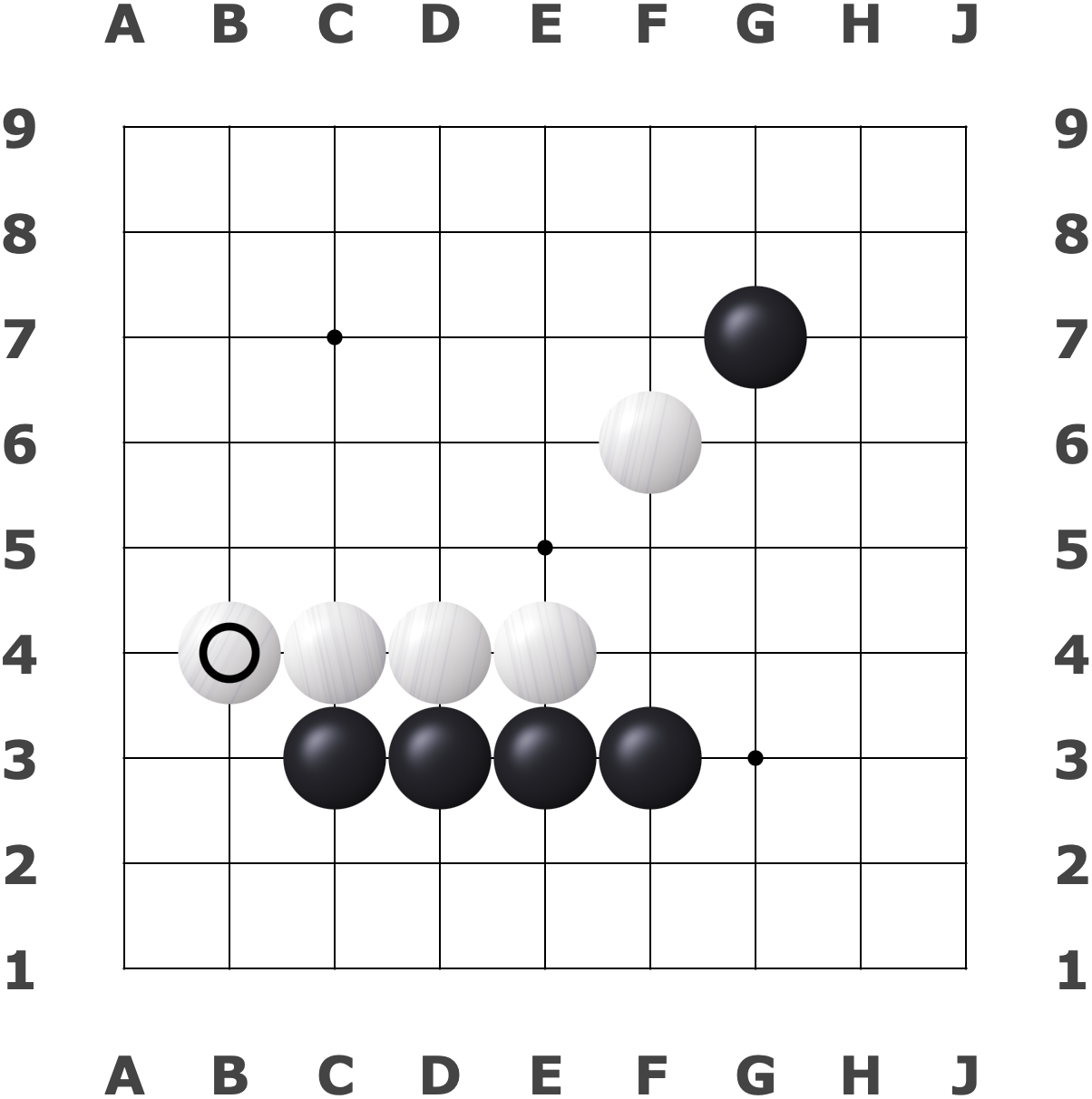}
  \caption{Wall}
  \label{fig:wall}
\end{subfigure}
\caption{Example Go patterns and associated terminology. (a) \textit{Cuts} are moves that separate two groups of stones. (b) \textit{Eyes} are empty squares surrounded by stones of the same color. (c) \textit{Ladders} are capturing races which may span the entire board. (d) Walls are lines of adjacent stones of the same color. These terms appear frequently in our dataset of natural language annotations and are further defined in the appendix.}
\label{fig:patterns}
\end{figure*}
\noindent Go is fundamentally a game of pattern recognition: from \textit{ladders} and \textit{walls} to \textit{sente} and \textit{shape}, professional players rely on a rich set of concepts to communicate about structures on the game board. 
Some patterns are relatively simple: \textit{walls} are lines of adjacent stones, and an \textit{atari} is a threat to capture stones on the next move; other patterns are less clearly defined: \textit{hane} refers to any move that ``goes around'' the opponent's stones, and \textit{sente} describes a general state of influence or tempo. 
Despite the nebulous definitions of some of these terms, human players use them productively. Beginners learn about \textit{eyes} that determine when groups of stones are \textit{alive} or \textit{dead} and are given guidelines for when they should play a \textit{cut} or extend a \textit{ladder}; more advanced players learn sequences of \textit{joseki} and \textit{tesuji} and are taught to distinguish \textit{good shape} from \textit{bad shape}. Figures~\ref{fig:comment}-\ref{fig:patterns} depict some example concepts.

Computers have recently surpassed human performance at Go \citep{silver2016mastering}, but relatively little is known about why these programs perform so well and whether they rely on similar representational units to choose the moves they play. While post-hoc behavioral analyses suggest that AlphaGo and its successor AlphaGo Zero \citep{silver2017mastering} can process complex game situations involving \textit{shape}, \textit{capturing races}, \textit{sente}, \textit{tesuji}, and even \textit{ladders}, existing interpretability work has focused on the moves that agents play, rather than the internal computations responsible for those moves. Our work instead proposes a \textit{structural} analysis by correlating the internal representations of game-playing agents with information from a naturally-occurring dataset of move-by-move annotations.

In this paper, we use linear probing to explore how domain-specific concepts are represented by game-playing agents. Because we do not have ground-truth labels explaining which concepts are relevant to a given game state, we collect a dataset of 10K annotated Go games (\S\ref{sec:games}).
Given a board state and its associated comment, we produce binary feature vectors summarizing which game phenomena (e.g., \textit{ko}, \textit{atari}) are mentioned in the comment and use pattern-based feature extractors to determine which phenomena are actually present on the board (\S\ref{sec:feats}).
We then feed board states into two policy networks with disparate architectures and training methods (\S\ref{sec:policy}) to obtain intermediate representations. Finally, we use linear probes (\S\ref{sec:probes}) to predict the binary feature vectors from our policy networks. Generally, we find that pattern-based features are encoded in the early layers of policy networks, while natural language features are most easily extracted from the later layers of both models. We release our code and data at \url{https://github.com/andrehe02/go}.

\section{Dataset}\label{sec:dataset}
\begin{figure*}[h]
\centering
\begin{subfigure}{0.45\textwidth}
\includegraphics[width=\linewidth]{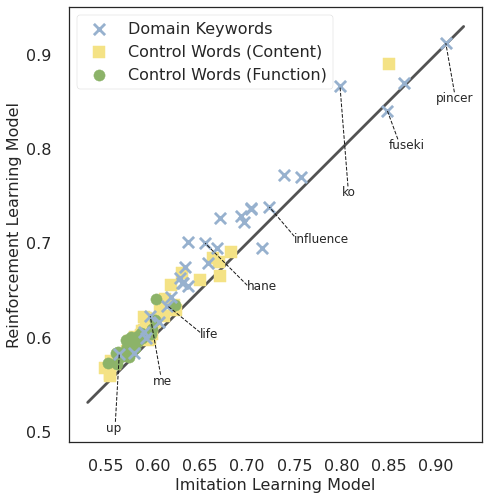}
\end{subfigure}
\begin{subfigure}{0.54\textwidth}
\includegraphics[width=\linewidth]{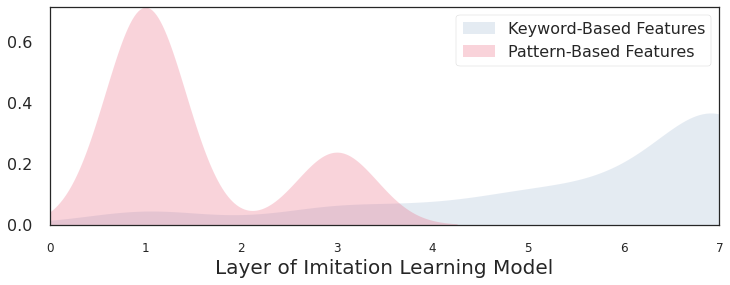} \\ \ \vspace{-0.4cm} \\
\includegraphics[width=\linewidth]{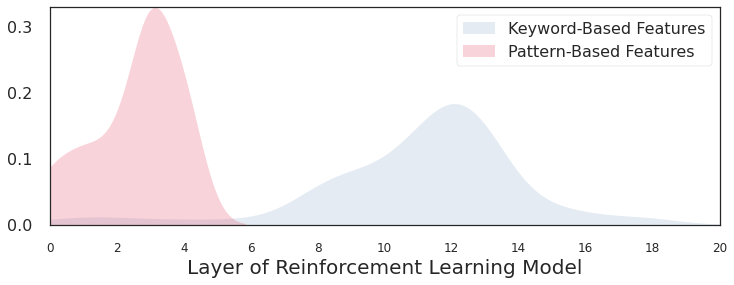}
\end{subfigure}
\caption{Results for the imitation learning and reinforcement learning agents are highly correlated. (Left) Scatterplot of ROC AUC values for linear probes trained to predict the presence of domain-specific keywords in move-by-move annotations. Keywords but not control words are predictable from the intermediate layers of both models. Words to the left of the solid line ($y=x$) are better predicted from the reinforcement learning model. (Right) Kernel density estimates showing where information is best represented in the policy networks (cf. {\S}\ref{sec:metrics}). For both policy networks, pattern-based features are encoded in early layers, while keyword-based features are most easily extracted from later layers. Layer 0 denotes the input board representation for both models.}
\label{fig:results}
\end{figure*}

\subsection{Annotated Games}\label{sec:games}
We collect 10K games with move-by-move English annotations from the Go Teaching Ladder (GTL).\footnote{\url{https://gtl.xmp.net}} The GTL was created by Jean-loup Gailly and Bill Hosken in 1994 and maintained until 2016 and permits non-commercial digital redistribution. Until 2016, members of the GTL could submit games for review by volunteers, who ranged from amateur to professional strength. Reviewers were given annotation guidelines and required to have a higher rating than their assigned reviewees, resulting in high quality natural language data. Of the collected games, we focus on 9524 which were played on classical $19 \times 19$ boards; many games also include unplayed analysis variations which we do not use in this work. These 9524 games contain 458,182 total comments, with a median length of 14 words.

\subsection{Feature Extraction}\label{sec:feats}
We convert board states and comments into binary feature vectors using two methods: (1) \textit{pattern-based} feature extraction, which checks for the ground truth presence of features from the board state, and (2) \textit{keyword-based} feature extraction, which converts comments into bag-of-words representations based on domain-specific keywords.

\paragraph{Pattern-Based} We define a set of rules to determine which game phenomena are present in a given board state, including: \textit{cuts}, \textit{eyes}, \textit{ladders}, and \textit{walls}. For example, we decide that a \textit{wall} is present when four stones of the same color are placed in a row adjacent to one another. Because patterns like \textit{wall} and \textit{cut} are often imprecisely defined, these definitions may not align perfectly with player intuitions; we therefore provide additional details for each phenomena in the appendix. We do not attempt to write rule-based definitions of vaguer concepts like \textit{sente} and \textit{influence}.

 \paragraph{Keyword-Based} We scrape an online vocabulary of domain-specific terminology\footnote{\url{https://senseis.xmp.net/?GoTerms}} and find the 30 most common terms in our natural language annotations. We then convert each comment into a 30-dimensional binary feature vector representing whether or not it contains these keywords; we additionally include features based on 60 control words, chosen according to frequency statistics, which are further subdivided into function and content words. Our wordlist and details about our selection of control words can be found in the appendix. 

\noindent Having both pattern-based and keyword-based features captures a trade-off between precision and coverage. Writing rules for pattern-based features is labor-intensive and essentially impossible for many game concepts. Meanwhile, keyword-based features are inherently noisy: comments often mention phenomena which didn't actually occur in the game, and common structures like \textit{atari} and \textit{eyes} are frequently left unmentioned because the annotator and players already know they exist.
Nonetheless, we find that probes are capable of predicting the presence of domain-specific keywords with significantly better-than-chance accuracy.

\section{Methods}
\subsection{Policy Networks}\label{sec:policy}
We analyze two agents: (1) an imitation learning agent using the architecture described in \citet{clark2015training}, and (2) a pre-trained ELF OpenGo model \citep{tian2017elf, tian2019elf}, which is an open-source, reinforcement learning agent similar to AlphaGo Zero \citep{silver2017mastering}. Our imitation learning model was trained on 228,000 games and achieved a rating of 1K ($\approx$ 1900 ELO) on the Online Go Server (OGS),\footnote{\url{https://online-go.com}} where it played against a combination of humans and computers until its rating stabilized. ELF OpenGo reports a self-play ELO over 5000, but this metric is inflated \citep{tian2019elf}. 
Although we refer to these agents by their training procedure (i.e., imitation vs. reinforcement), there are several other differences between the models. One possible source of variance between agents involves the format of the board state representation. Following \citet{clark2015training}, our imitation learning model takes as input a $19 \times 19 \times 7$ binary matrix. Of the seven planes, six represent the positions of stones, divided by color and the number of \textit{liberties}; the seventh plane represents \textit{ko} information. Meanwhile, the reinforcement learning model's $19 \times 19 \times 17$ input contains a partial history of the game state.

\subsection{Linear Probes}\label{sec:probes}
Given a board state and paired feature vector as described in Section~\ref{sec:feats}, we compute intermediate representations by feeding the board state into frozen policy networks. To predict each feature of interest, we run logistic regression independently on each layer of each policy network, including the raw board state.
In other words, for each policy network, we train $F \times L \times k$ classifiers, where $F$ is the number of features, $L$ is the number of layers, and $k$ is the parameter for $k$-fold cross-validation, as discussed in the following section.

\begin{table*}[h]
\centering
\begin{tabular}{lccl}
\toprule
Domain Word & Imitation & Reinforcement & Rough Definition \\
\midrule
\textit{Pincer} & 0.91 & 0.91 & attack on a corner approach \\
\textit{Joseki} & 0.87 & 0.87 & fixed local sequences of moves \\
\textit{Fuseki} & 0.85 & 0.84 & opening \\
\textit{Ko} & 0.80 & 0.86 & repetitive capture sequence \\
\textit{Wall} & 0.70 & 0.74 & sequence of stones in a row \\
\textit{Atari} & 0.69 & 0.73 & threat to capture \\
\textit{Eye} & 0.67 & 0.73 & surrounded empty space \\
\textit{Cut} & 0.64 & 0.65 & block two groups from connecting \\
\textit{Me} & 0.60 & 0.62 & another word for \textit{eye} \\
\textit{Down} & 0.60 & 0.60 & toward the edge of the board \\
\textit{Point} & 0.59 & 0.61 & specific locations on the board; or, the score \\
\textit{Force} & 0.58 & 0.58 & requiring immediate response \\
\textit{Up} & 0.56 & 0.58 & toward the center of the board \\
\bottomrule
\end{tabular}
\caption{ROC AUC values for a subset of domain words in both the imitation learning and reinforcement learning models. Higher values correspond to more predictable words. Domain words with the highest values represent relatively straightforward corner patterns (\textit{pincer}), while keywords with the lowest values (\textit{force}, \textit{up}) are polysemous with commonly used non-domain-specific meanings. See Table~\ref{tab:valuesfull} in the appendix for additional ROC AUC values.}
\label{tab:values}
\end{table*}

\subsection{Metrics}\label{sec:metrics}
We seek to answer two questions: (1) \textit{what} information is represented in the policy networks, and (2) \textit{where} is this information represented?
To answer the first question, we compute the area under the receiver operating characteristic curve (ROC AUC) for each linear probe. Specifically, for each layer, we compute the average ROC AUC after $10$-fold cross-validation and then take the maximum average value across layers. Features with high ROC AUC are said to be \textit{represented} by a model, because they are linearly extractible from some intermediate layer of its policy network. To answer the second question, we compute the layer at which each feature has its highest ROC AUC value; we then apply $10$-fold cross-validation, summarize the counts for each feature in a histogram, and compute a kernel density estimate (KDE) for visualization.

\section{Results}
We find that domain-specific keywords are significantly more predictable than control words, with $p = 1.8\times 10^{-5}$ under the Wilcoxon signed-rank test. As shown in Figure~\ref{fig:results} (Left) and Table~\ref{tab:values}, the keyword with the highest ROC AUC value across both models is \textit{pincer}, which denotes a relatively straightforward corner pattern. Meanwhile, low-valued domain words like \textit{me} and \textit{up} are polysemous with non-domain-specific meanings and therefore difficult to predict. While content and function control words have roughly similar distributions, some content words are noticeably more predictable; for example, \textit{opponents} is the highest-valued control word with ROC AUC values of $(0.85, 0.89)$ as seen in Figure~\ref{fig:results} (Left). Such control words are likely predictable due to correlations with certain domain-specific concepts.

ROC AUC values for the two models are strongly correlated, with Pearson's coefficient $\rho = 0.97$. Figure~\ref{fig:results} (Left) shows that for most keywords, the reinforcement learning model slightly outperforms the imitation learning model. Furthermore, keywords are significantly more predictable from the imitation learning model than from a randomly initialized baseline with identical architecture ($p=5.6\times10^{-16}$). Some words like \textit{ko} are noticeably more predictable from the reinforcement learning model, possibly due to differences in input board state representations (cf. \S\ref{sec:policy}); further discussion of this point can be found in the appendix.

Consistent with our knowledge that pattern-based features can be obtained by applying simple rules to the raw board state, we find that pattern-based features are encoded in early layers of both models, as shown in Figure~\ref{fig:results} (Right).
Meanwhile, keyword-based features are most easily extracted from later layers, suggesting that they correlate with high-level abstractions in the policy network.
Generally, pattern-based features are much more predictable than keyword-based features, with average ROC AUC values of $(0.96, 0.98)$ and $(0.68, 0.70)$, respectively. As discussed in Section~\ref{sec:feats}, this discrepancy can largely be attributed to the noisiness inherent in natural language data.

\section{Related Work}
\citet{jhamtani2018learning} propose a similarly-sized dataset of move-by-move chess commentary. Rather than using this commentary for model interpretability, though, \citet{jhamtani2018learning} attempt to predict whole comments from raw board states. \citet{zang2019automated} use the same dataset to jointly train a policy network and language generation model with a shared neural encoder, but again focus on the pedagogical application of commentary generation rather than interpretation of the policy network. Similar work has focused on generating sportscasts in the Robocup domain \citep{chen2008learning, liang2009learning, mei2016talk}.

Our primary methodology is linear probing \citep{ettinger2016probing, manning2020emergent}, which has commonly been used to study the intermediate representations of language models like ELMo \citep{peters2018deep} and BERT \citep{devlin2019bert}. One classic result in this area shows that early layers of contextual language models correlate best with lexical-syntactic information such as part of speech, while later layers correlate with semantic information like proto-roles and coreference \citep{tenney2019bert}. Recent work on control tasks \citep{hewitt2019designing}, minimum description length \citep{voita2020information}, and Pareto probing \citep{pimentel2020pareto} has focused on improving the methodological rigor of this paradigm. Although linear probing is fundamentally a correlational method, other recent work has focused on whether information which is easily extractable from intermediate layers of a deep network is causally used during inference \citep{elazar2021amnesic, lovering2021predicting}.

Most related to our work are contemporary studies by \citet{mcgrath2021acquisition} and \citet{forde2022hex}, which apply probing techniques to the games of chess and Hex, respectively. \citet{mcgrath2021acquisition} use linear probes to predict a large number of pattern-based features throughout the training of an AlphaZero agent for chess. Meanwhile, \citet{forde2022hex} train linear probes for pattern-based features on an AlphaZero agent for Hex and run behavioral tests to measure whether the agent ``understands'' these concepts. Comparatively, our work uses fewer features than \citet{mcgrath2021acquisition} and does not make causal claims about how representations are used during inference, as in \citet{forde2022hex}; however, to the best of our knowledge, our work is the first of its kind to use features derived from natural language in conjunction with probing techniques for policy interpretability.

\section{Conclusion}
We presented a new dataset of move-by-move annotations for the game of Go and showed how it can be used to interpret game-playing agents via linear probes.
We observed large differences in the predictability of pattern-based features, which are extracted from the board state, and keyword-based features, which are extracted from comments.
In particular, pattern-based features were easily extracted from lower layers of the policy networks we studied, while keyword-based features were most predictable from later layers.
At a high level, this finding reinforces the intuition that written annotations describe high-level, abstract patterns that cannot easily be described by a rule-based approach. Accordingly, we argue there is much to learn from this annotation data: future work might attempt to correlate policy network representations with richer representations of language, such as those provided by a large language model. Future work might also explore whether similar approaches could be used to improve game-playing agents, either by exposing their weaknesses or providing an auxiliary training signal. We also expect similar approaches may be viable in other reinforcement learning domains with existing natural language data.

\section*{Acknowledgements}
We are grateful to Kevin Yang for his work on the imitation learning model, and to Rodolfo Corona and Ruiqi Zhong for their contributions to an early version of this project. We thank Roma Patel, the members of the Berkeley NLP Group, and our anonymous reviewers for helpful suggestions and feedback. This work was supported by the DARPA XAI and LwLL programs and a National Science Foundation Graduate Research Fellowship.

\bibliographystyle{acl_natbib}
\bibliography{references}

\clearpage
\appendix
\section{Dataset Statistics}
\begin{figure}[h]
\centering
\includegraphics[width=\linewidth]{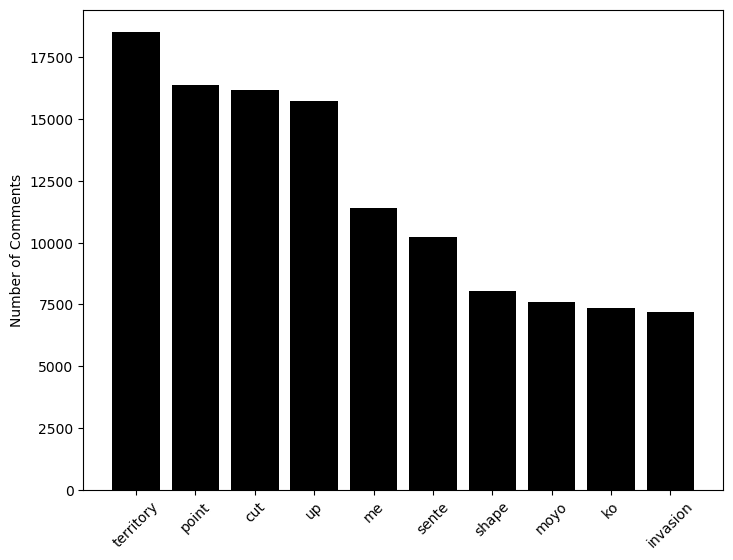}
\caption{Histogram showing the number of comments that contain each of the ten most common domain-specific words.}\label{fig:histogram}
\end{figure}

The most common domain-specific terms appear in more than 15K comments, as shown in Figure~\ref{fig:histogram}.

\section{Pattern-Based Feature Extraction}\label{sec:featdetails}
As described in Section~\ref{sec:feats}, we extract features from the board state using a set of hand-crafted rules. These rules may not align perfectly with player intuitions, which can be hard to formulate concisely, and so are presented here for full detail.

\paragraph{Cut}
We define \textit{cuts} as moves that prevent the opponent from connecting two disconnected groups on their next move. To avoid labelling squares where a play would be immediately capturable, we also require that a cut have at least two liberties. Note that this definition permits non-diagonal cuts.

\paragraph{Eye} We define \textit{eyes} as connected groups of empty squares that are completely surrounded by stones of the same color. We require there be no enemy stones in the same surrounded region, so this definition fails to capture eyes that surround \textit{dead} stones.

\paragraph{Ladder} The term \textit{ladder} describes the formation shown in Figure~\ref{fig:ladder}. Since human players can usually predict who wins a ladder, they rarely play out the capturing race. For this reason, we do not look for ladder formations, but instead label moves that would start or continue a ladder. Specifically, we label a square for the ladder feature if it is the singular liberty of a friendly group of stones and a play at the square results in the group having exactly two liberties.
We do not count trivial ladders that lie at the edge of the board.

\paragraph{Wall} We define a \textit{wall} as a connected row or column with four or more stones of the same color.

\section{Keywords and Control Words}
\paragraph{Keywords} We choose the first thirty most frequent terms (cf. Table~\ref{tab:values}) from our vocabulary of domain-specific terminology as keywords: \textit{territory, point, cut, sente, up, me, moyo, shape, ko, invasion, influence, wall, joseki, eye, alive, gote, life, pincer, aji, thickness, base, atari, connected, hane, tenuki, down, overplay, force, reading, fuseki}.

\paragraph{Control Words} Our control words consist of the thirty most frequent words in our dataset, as well as thirty words uniformly distributed according to the same frequency as the keywords: \textit{the, is, to, this, white, black, and, you, at, for, in, move, it, of, but, not, be, have, play, that, on, good, here, if, better, can, would, now, should, stones, looking, wanted, opponents, wasnt, defending, save, youre, answer, three, fine, feel, place, lose, bit, possibility, attacking, likely, leaves, shouldnt, question, lost, threat, almost, theres, continue, trying, hope, just, exchange, before}. We further subdivide the control words based on whether or not they appear in the NLTK stopword list,\footnote{\url{https://gist.github.com/sebleier/554280}} which we use as a rough proxy for distinguishing between function and content words.

\section{Additional Results}
\begin{figure*}[h]
\centering
\includegraphics[width=\textwidth]{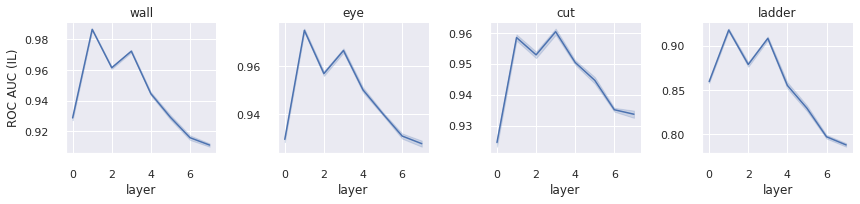}
\includegraphics[width=\textwidth]{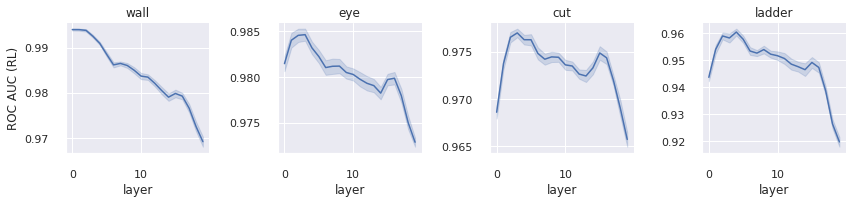}
\caption{Plots of ROC AUC values for pattern-based features in the imitation learning model (top) and the reinforcment learning model (bottom). Among the four pattern-based features we consider, \textit{ladders} have by far the lowest ROC AUC values. As noted in \citet{tian2019elf}, \textit{ladders} are a known challenge for Go agents, requiring special feature engineering in \citet{silver2016mastering}. Therefore, perhaps it is unsurprising that \textit{ladders} were the most difficult pattern-based feature to predict.}
\label{fig:ilpattern}
\end{figure*}

\begin{figure*}[h]
\centering
\includegraphics[width=\textwidth]{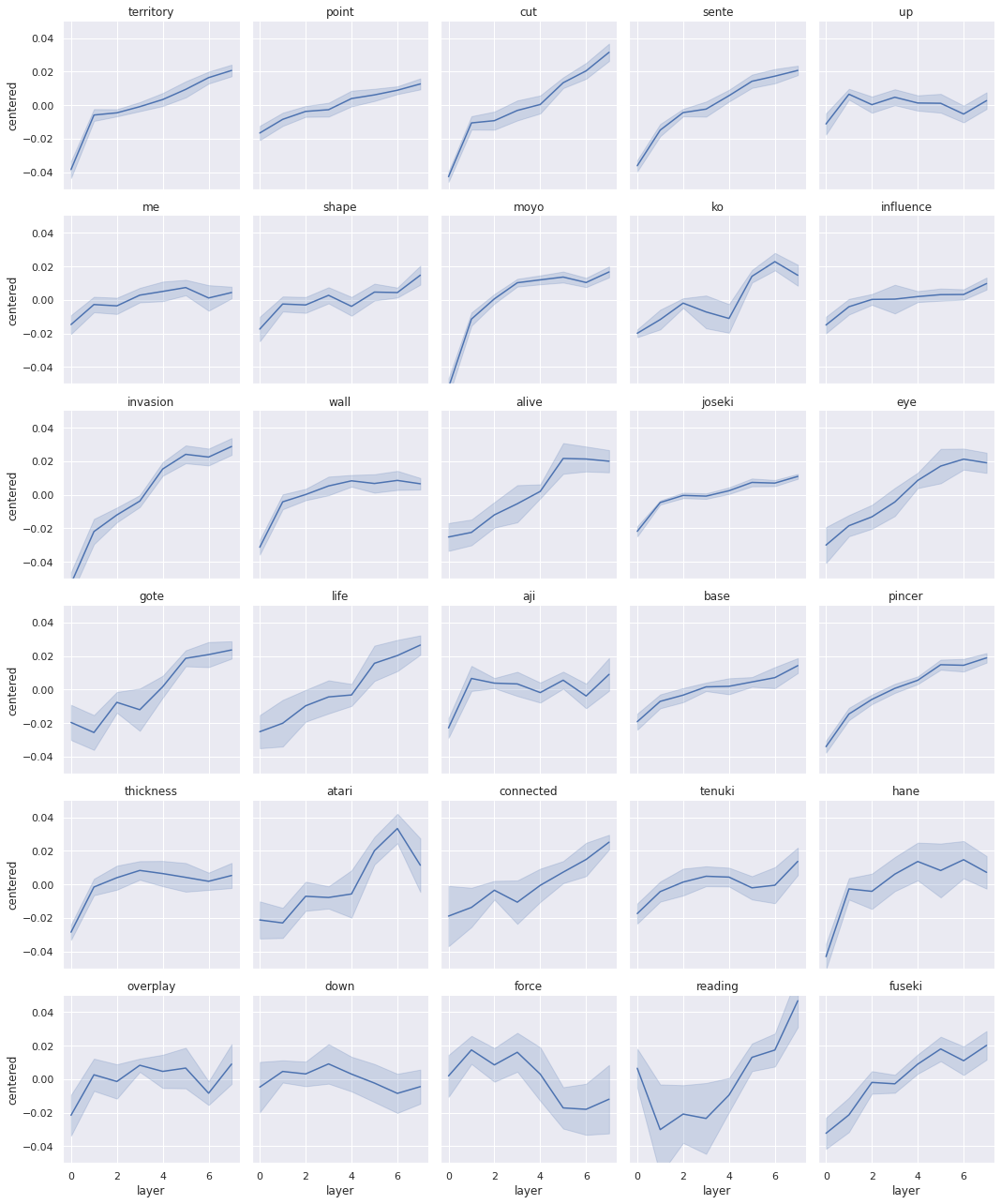}
\caption{Plots of centered ROC AUC values for keywords in the imitation learning model. For most keywords, linear probe performance peaks at mid-to-late layers of the imitation learning model.}
\label{fig:ilraw}
\end{figure*}

\begin{figure*}[h]
\centering
\includegraphics[width=\textwidth]{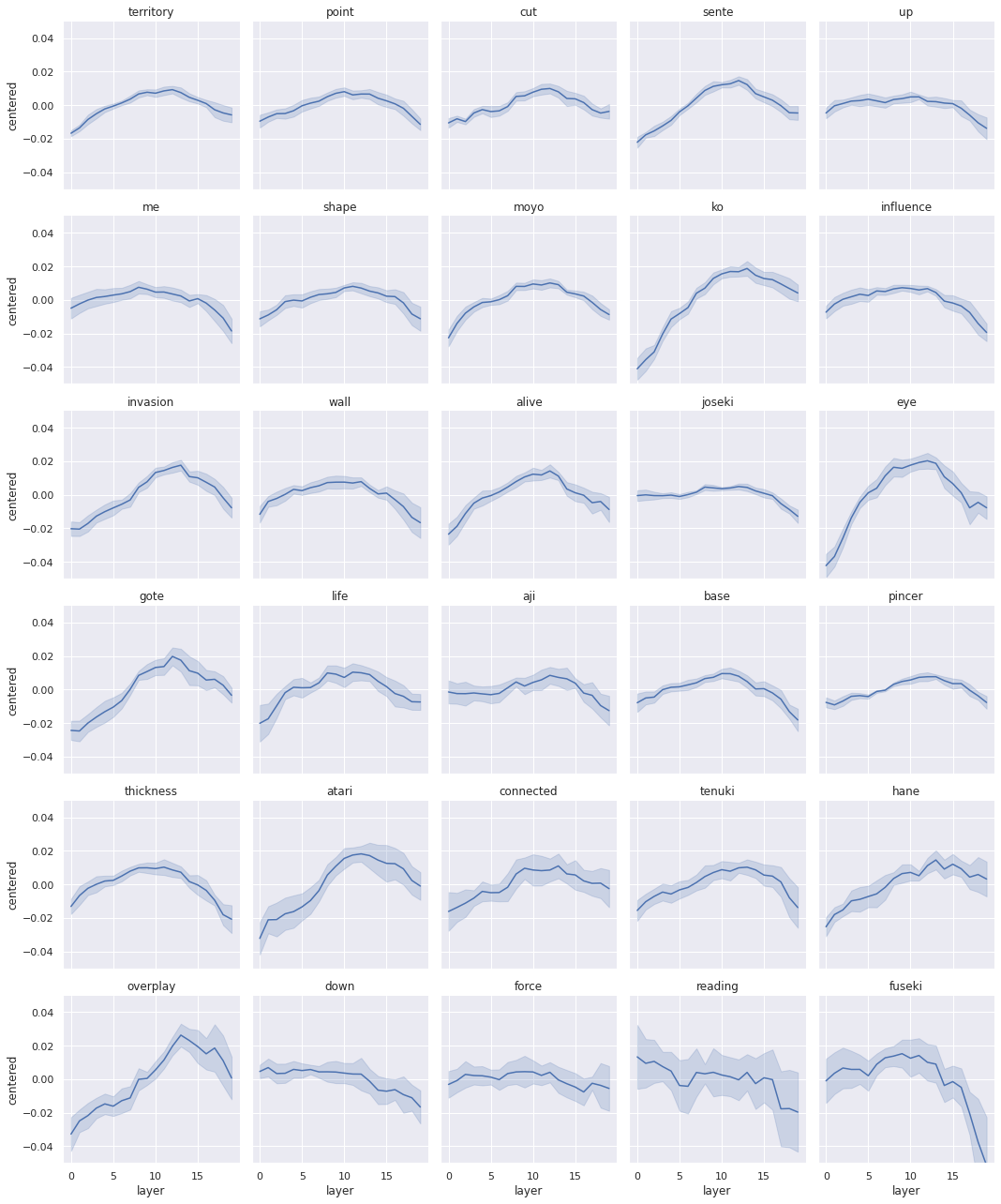}
\caption{Plots of centered ROC AUC values for keywords in the reinforcement learning (RL) model. While noisier than the imitation learning model, probe performance still tends to peak at mid-to-late layers of the RL model and declines at the final layers. Figure~\ref{fig:results} aggregates these results alongside pattern-based feature classifiers.}
\label{fig:elfraw}
\end{figure*}

We additionally report de-aggregated ROC AUC values for each keyword across layers, as shown in Figures~\ref{fig:ilpattern}-\ref{fig:elfraw}. These figures show the raw data used to compute the kernel density estimates in Figure~\ref{fig:results}, which show that natural language features are most easily extracted from later layers of both models. We note in Figure~\ref{fig:ilpattern} that \textit{ladders} are the most difficult pattern-based feature to predict, which is consistent with our knowledge that many Go-playing agents fail to correctly handle ladders without special feature engineering \citep{tian2019elf}. Anecdotally, our imitation learning model often failed to play ladders correctly; this is consistent with the finding that ladders are more predictable from the reinforcement learning model. Future work might investigate whether this probing framework could be used to effectively predict model behavior in situations like these, as in \citet{forde2022hex} for the game of Hex. 

\section{Major Differences Between Imitation and Reinforcement Learning Models}\label{sec:qual}
While most keywords have similar ROC AUC values across models, \textit{ko}, \textit{eye}, \textit{atari}, and \textit{overplay} have a noticeably higher ROC AUC values under the reinforcement learning model (cf. Table~\ref{tab:values}). However, this discrepancy is not obviously attributable to the difference in training procedures (i.e., imitation vs. reinforcement). As described in Section~\ref{sec:policy}, the two models use different input state representations, which differ in their encoding of \textit{ko} and \textit{liberty} information, which is used to determine whether \textit{eyes} and \textit{atari} exist. Such architectural differences may explain discrepancies across models, but do not account for words like \textit{overplay}; playing strength is another possible (but not confirmed) source of these discrepancies.

\begin{table*}[b]
\centering
\begin{tabular}{lccl}
\toprule
Domain Word & Imitation & Reinforcement & Rough Definition \\
\midrule
\textit{Pincer} & 0.91 & 0.91 & attack on a corner approach \\
\textit{Joseki} & 0.87 & 0.87 & fixed local sequences of moves \\
\textit{Fuseki} & 0.85 & 0.84 & opening \\
\textit{Ko} & 0.80 & 0.86 & repetitive capture sequence \\
\textit{Base} & 0.76 & 0.77 & starter eye space \\
\textit{Moyo} & 0.74 & 0.77 & sphere of influence \\
\textit{Influence} & 0.72 & 0.74 & long-range effect of stones \\
\textit{Reading} & 0.72 & 0.70 & calculating an upcoming sequence \\
\textit{Wall} & 0.70 & 0.74 & sequence of stones in a row \\
\textit{Thickness} & 0.70 & 0.74 & strength of a group of stones \\
\textit{Invasion} & 0.70 & 0.72 & attack on enemy territory \\
\textit{Atari} & 0.69 & 0.73 & threat to capture \\
\textit{Eye} & 0.67 & 0.73 & surrounded empty space \\
\textit{Gote} & 0.67 & 0.69 & loss of initiative \\
\textit{Tenuki} & 0.66 & 0.68 & non-local response \\
\textit{Hane} & 0.66 & 0.70 & move that ``reaches around'' or bends \\
\textit{Overplay} & 0.64 & 0.70 & overly aggressive move \\
\textit{Cut} & 0.64 & 0.65 & block two groups from connecting \\
\textit{Alive} & 0.63 & 0.67 & cannot be captured \\
\textit{Territory} & 0.63 & 0.66 & controlled empty space \\
\textit{Aji} & 0.63 & 0.66 & possibilities left in a position \\
\textit{Sente} & 0.63 & 0.66 & initiative \\
\textit{Shape} & 0.62 & 0.64 & quality of a group of stones \\
\textit{Life} & 0.62 & 0.63 & inability to be captured \\
\textit{Connected} & 0.61 & 0.62 & adjacent or nearby stones \\
\textit{Me} & 0.60 & 0.62 & another word for \textit{eye} \\
\textit{Down} & 0.60 & 0.60 & toward the edge of the board \\
\textit{Point} & 0.59 & 0.61 & specific locations on the board; or, the score \\
\textit{Force} & 0.58 & 0.58 & requiring immediate response \\
\textit{Up} & 0.56 & 0.58 & toward the center of the board \\
\bottomrule
\end{tabular}
\caption{ROC AUC values for all domain words in both the imitation learning and reinforcement learning models. Domain words with the highest values represent relatively straightforward corner patterns (\textit{pincer}), while keywords with the lowest values (\textit{force}, \textit{up}) are polysemous with commonly used non-domain-specific meanings.}
\label{tab:valuesfull}
\end{table*}

\end{document}